\documentclass[11pt]{article}
\pdfoutput=1
\usepackage{coling2020}
\usepackage{hyperref}
\usepackage{times}
\usepackage{url}
\usepackage{latexsym}
\usepackage{graphicx}
\usepackage{subcaption}
\usepackage{chngcntr}
\usepackage[titletoc,title]{appendix}

\usepackage{lscape}
\usepackage{rotating}
\colingfinalcopy 

\title{What is SemEval evaluating? \\ A Systematic Analysis of Evaluation Campaigns in NLP}

\author{Oskar Wysocki, Malina Florea \and Andre Freitas \\
         Department of Computer Science \\ The University of Manchester,  United Kingdom \\ 
         {\tt oskar.wysocki@manchester.ac.uk} \\
  {\tt malina.florea@student.manchester.ac.uk} \\
  {\tt andre.freitas@manchester.ac.uk}}

\date{}

\begin{document}
\maketitle
\begin{abstract}
  SemEval is the primary venue in the NLP community for the proposal of new challenges and for the systematic empirical evaluation of NLP systems. This paper provides a systematic quantitative analysis of SemEval aiming to evidence the patterns of the contributions behind SemEval. By understanding the distribution of task types, metrics, architectures, participation and citations over time we aim to answer the question on what is being evaluated by SemEval.
\end{abstract}

\section{Introduction}

A large portion of the empirical methods in Natural Language Processing (NLP) are defined over canonical text interpretation tasks such as Named Entity Recognition (NER), Semantic Role Labeling (SRL), Sentiment Analysis (SA), among others. The systematic creation of benchmarks and the comparative performance analysis of resources, representations and algorithms is instrumental for moving the boundaries of natural language interpretation. SemEval \cite{semeval-2019-international,semeval-2018-international,semeval-2017-international,semeval-2016-international,semeval-2015-international,semeval-2014-international,semeval-2013-joint-lexical,semeval-2012-sem} is the primary venue in the NLP community for the organisation of shared NLP tasks and challenges. SemEval is organised as an annual workshop co-located with the main NLP conferences and has attracted a large and growing audience of task organisers and participants. 

Despite its recognition as a major driver in the creation of gold-standards and evaluation campaigns, there is no existing meta-analysis which interprets the overall contribution of SemEval as a collective effort. This paper aims to address this gap by performing a systematic descriptive quantitative analysis of 96 tasks encompassing the SemEval campaigns between 2012-2019. This study targets understanding the evolution of SemEval over this period, describing the core patterns with regard to task popularity, impact, task format (inputs, outputs), techniques, target languages and evaluation metrics. 

This paper is organised as follows: section 2 describes related work; 3 describes the methodology; 4 defines the underlying task macro-categories; 5 and 6 presents the number of tasks and popularity in 2012-2019; 7 discusses SemEval impact in terms of citations; 8 shows targeted languages; then, sections 9, 10, 11 analyse input, output and evaluation metrics; 11 focuses on sentiment analysis architectures and representations; this is followed by a Discussion section; we close the paper with Recommendations and Conclusions.

\section{Related work}
Each SemEval task is described by an \textit{anthology}, which contains: a summary of previous editions or similar tasks, references to previous works, detailed task description, evaluation methods, available resources, overview of submitted systems and final results of the competition. It is worth noting, there is a variation, or even inconsistency, in the structure and the level of detail in the description. Participants are also encouraged to submit papers with systems architecture explanations. However, there is a lack of overall analysis across different tasks and years in SemEval. There are existing studies on the analysis of specific SemEval tasks. \cite{Nakov2016} focuses on developing Sentiment Analysis tasks in 2013-2015. \cite{Sygkounas2016} is an example of a replication study of the top performing systems, in this case systems used in SemEval Twitter Sentiment Analysis (2013-2015), and focuses on architectures and performance. Evolution and challenges in semantics similarity were described in \cite{JIMENEZ2015}. This is an example of a study on the performance of a given type of architecture across tasks of the same type. There also exist studies on shared tasks in given domain, specially in clinical application of NLP \cite{PMID:30157522}, \cite{10.1136/amiajnl-2011-000465}. However, they refer to tasks outside the SemEval and are more result oriented rather than task organization. Some studies discuss ethical issues in the organisation and participation of shared tasks. An overview focusing on task competitive nature and fairness can be found in \cite{parra-escartin-etal-2017-ethical}. In \cite{nissim-etal-2017-last} authors also relate to these issues, yet giving the priority to advancing the field over fair competition. 

Comparatively, this paper covers a wider range of NLP topics, and compares sentiment analysis and semantic similarity as well as other task types/groups in a systematic manner. To the best to our knowledge this is the first systematic analysis on SemEval.

\section{Analysis methodology}

We build a corpus based on the ACL anthology archive from the SemEval workshops between the years 2012-2019. Reference material included ACL anthology papers covering the task description, tasks' websites and papers describing the participating systems. All the reference papers included in this analysis are reported at the Appendix B. The pre-processing analysis consisted in manually extracting the target categories for the analysis which includes: task types, input and output types, as well as evaluation metrics, number of teams, languages and system architectures. Tasks were grouped based on the similarity between task types similarity. If the same team took part in several tasks the same year, we considered each participation as distinct. There are four missing tasks in the plotted indexes, due to cancellation (2015-16, 2019-11), task-sharing (2013-6) or lack of supporting task description (2013-14). Numbers of citations are the numbers returned by Google Scholar, using \textit{Publish and Perish} supporting API \cite{publish-and-perish}. The list of citations were manually validated and noisy entries were filtered out. A final table with all the values extracted from the corpus is included in the Appendix B.

\begin{figure*}[ht]
    \centering
    \begin{subfigure}[b]{0.475\textwidth}
        \centering
        \includegraphics[width=0.93\textwidth]{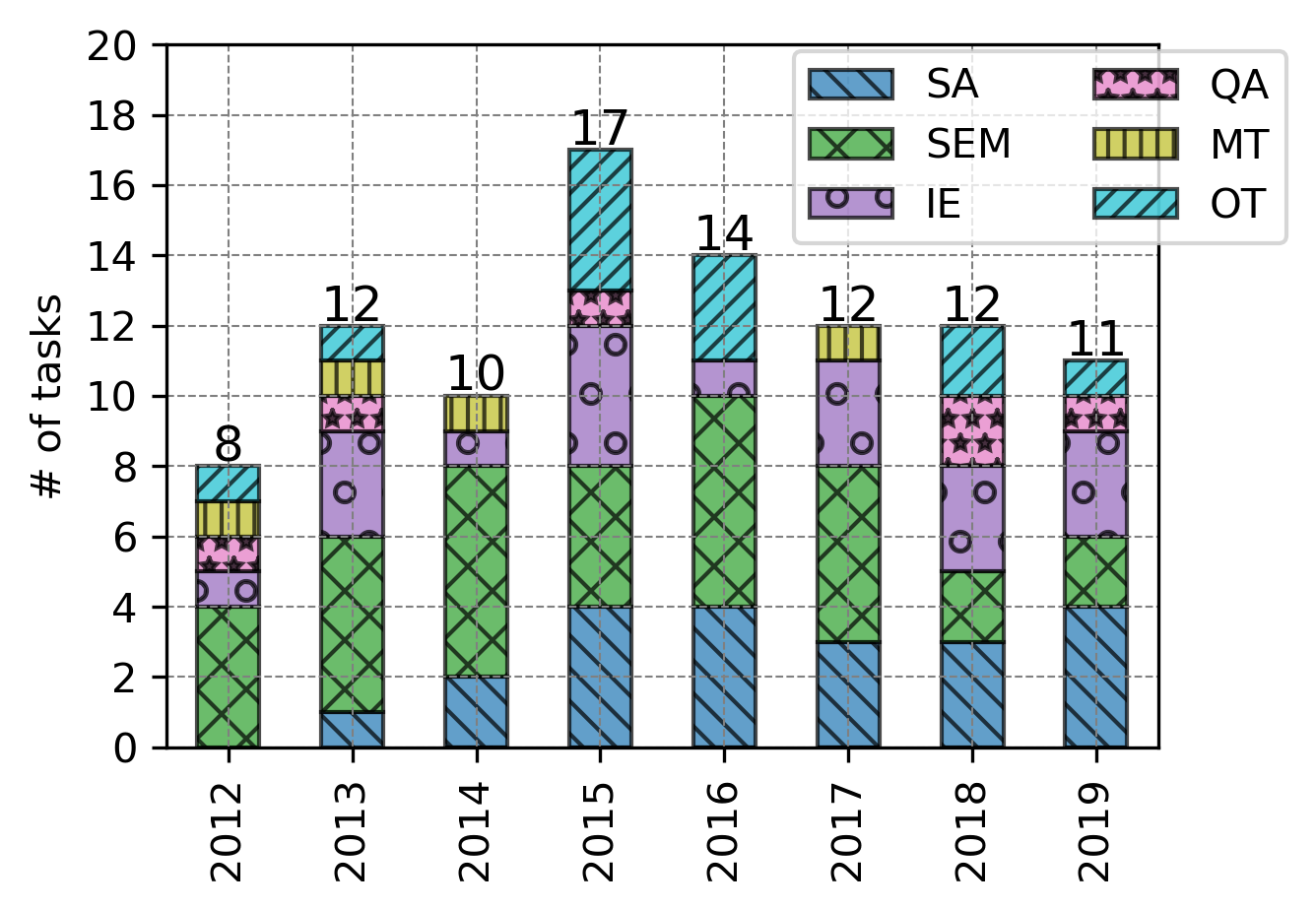}
        \caption[]%
        {{\small Number of tasks in SemEval 2012-2019}}    
        \label{fig:tasks_in_years}
    \end{subfigure}
    \hfill
    \begin{subfigure}[b]{0.475\textwidth}  
        \centering 
        \includegraphics[width=0.9\textwidth]{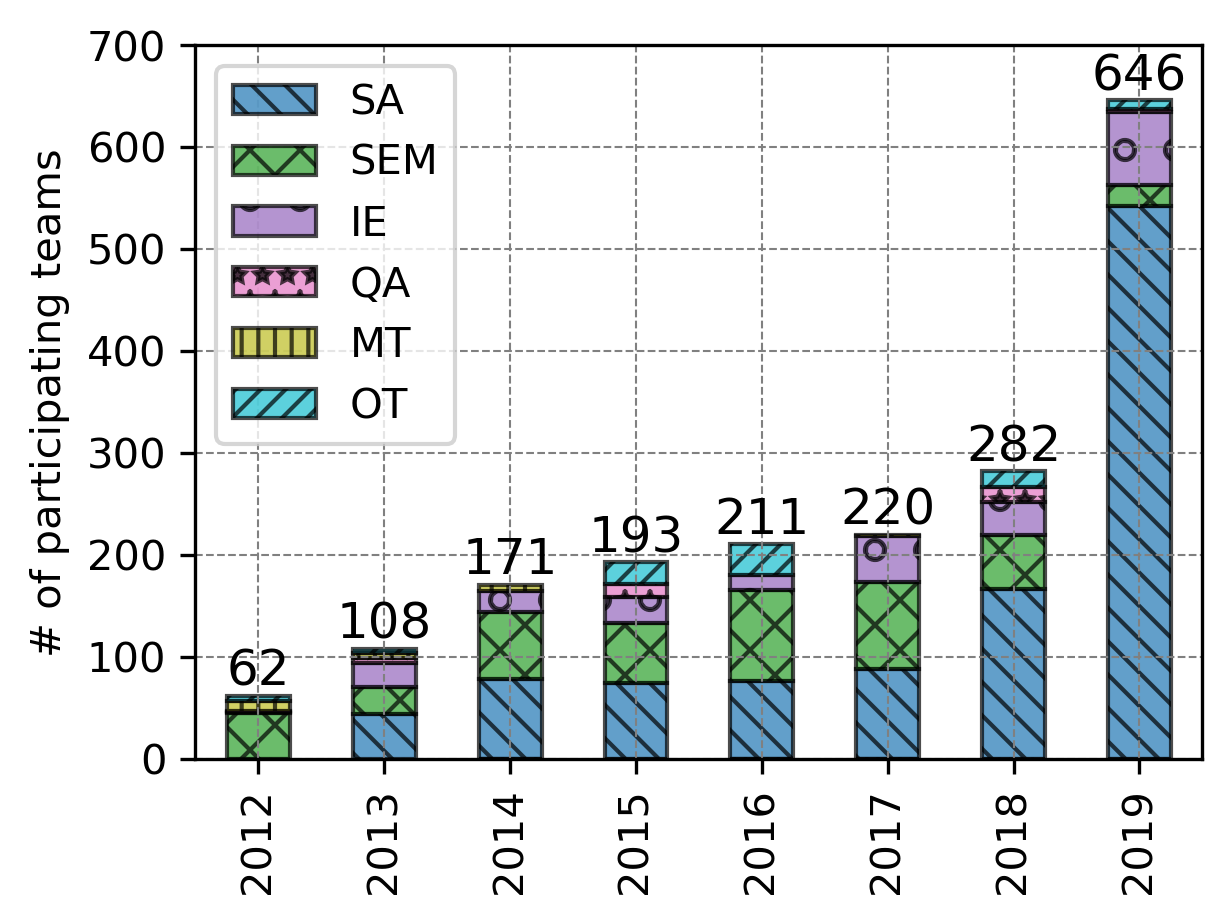}
        \caption[]%
        {{\small Number of teams participating in SemEval}}    
        \label{fig:teams}
    \end{subfigure}
    
    \vskip\baselineskip
    \begin{subfigure}[b]{0.475\textwidth}   
        \centering 
        \includegraphics[width=0.9\textwidth]{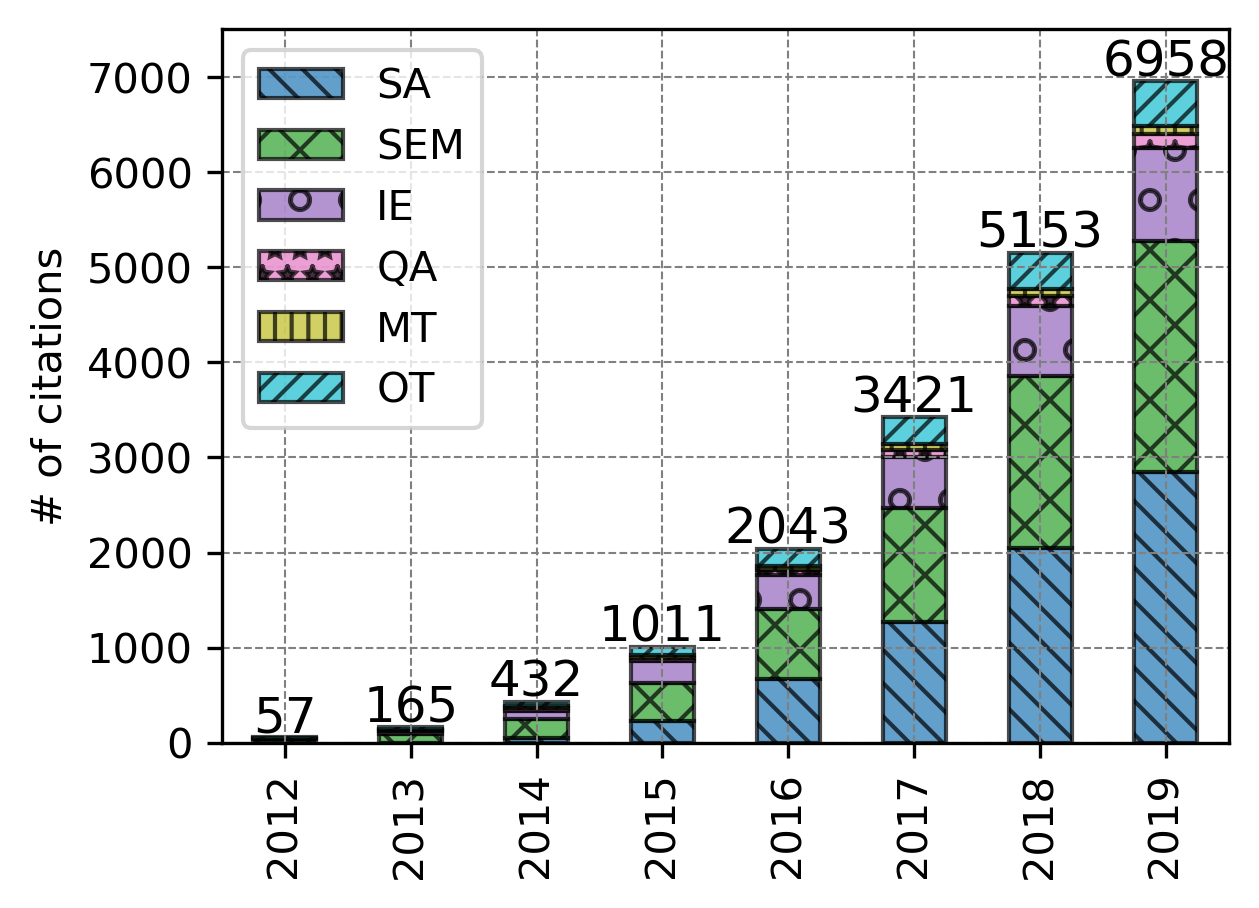}
        \caption[]%
        {{\small Cumulative number of task citations, except
for citations in SemEval proceedings}}    
        \label{fig:cum}
    \end{subfigure}
    \quad
    \begin{subfigure}[b]{0.475\textwidth}   
        \centering 
        \includegraphics[width=0.9\textwidth]{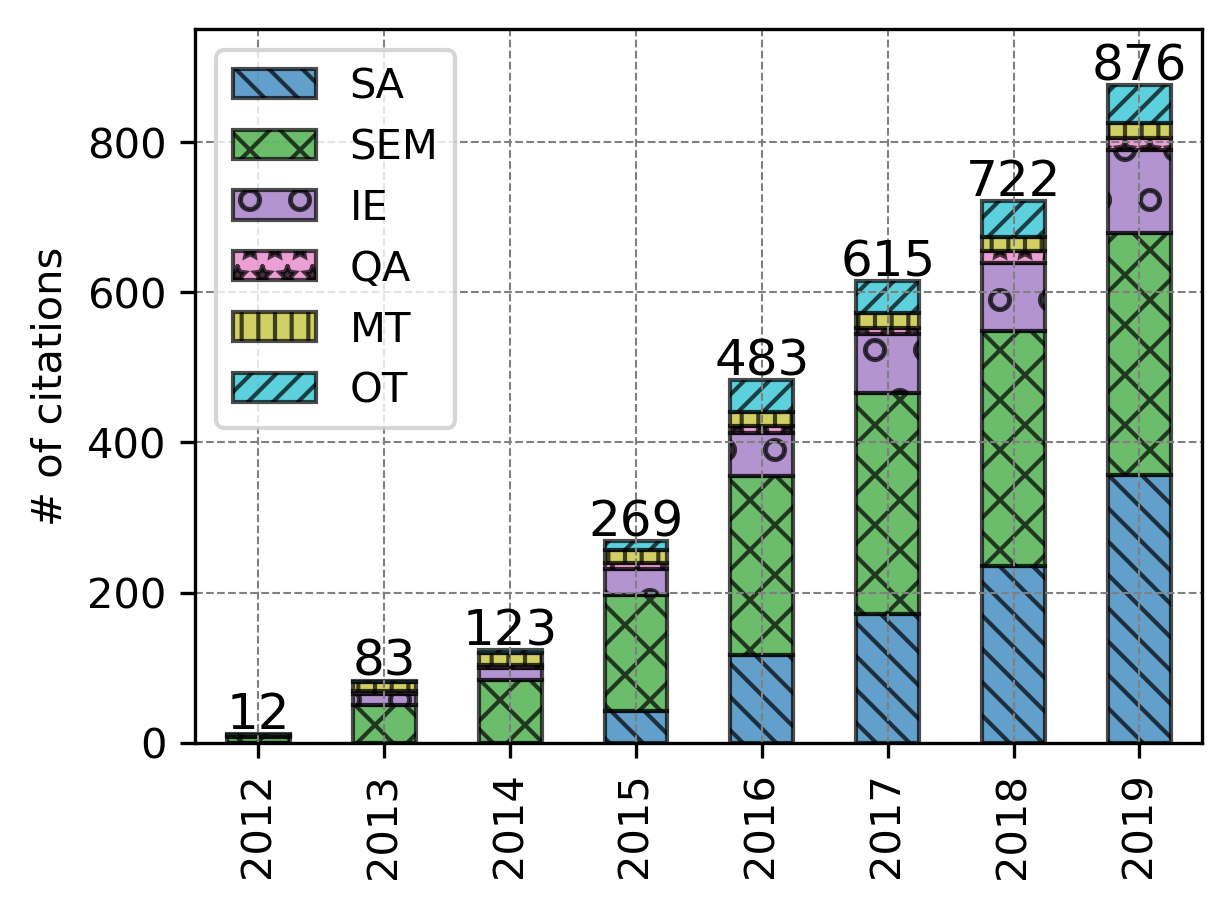}
        \caption[]%
        {{\small Cumulative number of task citations in SemEval proceedings}}    
        \label{fig:cum_int}
    \end{subfigure}
        
    \caption[ AAA ]
    {\small Overall plots for SemEval editions 2012-2019} 
    \label{fig:overall_plots}
\end{figure*}

\section{Task types and groups}
Based on task description we group each task within a macro-category. Then, due to a large number of task types, tasks were clustered within 6 groups: \textit{Sentiment Analysis} (\textbf{SA}); \textit{Semantic Analysis} (\textbf{SEM}): semantic analysis, semantic difference, semantic inference, semantic role labeling, semantic parsing, semantic similarity, relational similarity; \textit{Information Extraction} (\textbf{IE}): information extraction, temporal information extraction, argument mining, fact checking; \textit{Machine Translation} (\textbf{MT}); \textit{Question Answering} (\textbf{QA}); \textit{Other} (\textbf{OT}): hypernym discovery, entity linking, lexical simplification, word sense disambiguation, taxonomy extraction, taxonomy enrichment. There are also macro-categories defined by the SemEval organizers, starting from 2015, but we found them not consistent enough for the purpose of this analysis.

\section{SemEval tasks in years}
Within 8 editions of SemEval, a total of 96 tasks were successfully announced. The number of tasks within one group is roughly similar every year (except for MT), as well as distribution of tasks in each edition. According to Fig.\ref{fig:tasks_in_years}, we observe decreasing number of SEM tasks: 5 on average in 2012-2017, and only 2 in 2018-2019. Moreover, there were no machine translation tasks in the last 2 years, and a low number of MT tasks in general (only 4 tasks in 8 years). 

Although SA has a relatively limited task complexity when compared to SEM or IE, which reflects a higher variety of task types and an abundance of more specific interpretation challenges, the number of SA tasks each year is high (4, 3, 3 and 4 in years 2016-2019). It is worth mentioning, that there are other 6 SA tasks in the forthcoming SemEval 2020. The absence of some task types may be caused by the emergence of specialized workshops or conferences, e.g. low number of MT tasks in SemEval is caused by the presence a separate venue for MT: the Conference On Machine Translation \cite{barrault-EtAl:2019:WMT}, which attracts more participants than SemEval in this field.

\section{Task popularity}
As a measure of task popularity, we analysed how many teams participated in a given task. As the number of teams signed up to the task is usually much higher than the number submitting a system result, we consider only the latter. 

The number of teams increased significantly from 62 in 2012 to 646 in 2019, which shows not only a popularity boost for SemEval, but an increase in the general interest for NLP. So far, a total of 1883 teams participated in this period.

In Fig.\ref{fig:teams}, we observe a gradual increase in SemEval popularity, 30\% on average each year to 2018, with a +129\% jump in 2019. This is associated mainly with a dramatic increase of interest for SA: 542 teams (84\% of total) in 2019. However, at the same time, number of teams in non-SA tasks decreased from 132 in 2017, to 115 in 2018 and 104 in 2019.

The most popular tasks groups along the years are SA and SEM, which gather more than 75\% of teams on average each year. The third most popular is IE, in which total of 235 teams participated in SemEval from 2012 (12\% of total). As a contrast, we observe a relatively low interest in QA and OT tasks. Only 41 teams participated in the last 3 years (3\% of a total of 1148 in 2017-2019). Especially in OT tasks, which concentrates novel tasks, in many cases including novel formats. 

In the last 2 years, SA shows a major increase in popularity (76\% of all teams, compared to 40\% in 2013-2017). At the same time, in tasks such as 2019-10, 2018-4 and 2018-6, which are mathematical question answering, entity linking on multiparty dialogues and parsing time normalization, respectively, only 3, 4 and 1 teams submitted results. This divergence may be associated with an emergence of easily applicable ML systems and libraries, which better fit to standard classification tasks more prevalent in SA (in contrast to OT, QA nor IE).

\section{The impact of SemEval papers}

As a measure of the impact of SemEval outcomes in the NLP community, we analysed the numbers of citations per task description in Google Scholar. The task description paper was used as a proxy to analyse the task impact within the NLP community. Papers submitted by participating teams describing systems and methods were not included on this analysis.

We considered the cumulative citations from 2012 to 2019 (Fig.\ref{fig:cum}), with additional distinction on citations of task description papers published in a given year (Fig.\ref{fig:cit_year}). Citations within SemEval proceedings were treated separately, as we focused on the impact both outside (Fig.\ref{fig:cum}) and inside (Fig.\ref{fig:cum_int}) the SemEval community. In other words, citations found in Google Scholar are split into numbers of papers \textit{out} and \textit{in} the SemEval proceedings.

SA and SEM have the highest impact, being the most cited tasks along the years both inside and outside SemEval community, what can be attributed to their high popularity. 

Considering the external impact, in 2019 SA and SEM anthologies contributed with 2847 (41\%) and 2426 (35\%) citations respectively. IE has 985 citations (14\%) and QA contributed with 148 citations (2\%). The OT group, which consists of less canonical tasks, accumulated 468 citations (7\%). The impact of MT papers is noticeably lower - 84 (1\%).

In terms of citations within the SemEval community (in all SemEval 2012-2019 proceedings), we observe a similar pattern: 41\% and 37\% citations in 2019 come from SA and SEM (357 and 322), and for remaining task groups proportions are almost identical as in citations outside community (Chi.sq. \textit{p-value}=0.06).

The number of citations outside is 8 times higher than inside the community. This proves the scientific impact and coverage, which leads to beneficial effect of SemEval on the overall NLP community.

A total of 6958 citations from 2019 are depicted in Fig.\ref{fig:cit_year} with distinction on the year in which the task was published (e.g. tasks from 2016 are cited 1682 times (23\%)). Similarly, a total of 876 citations in the SemEval proceedings are presented in Fig.\ref{fig:cit_int} (e.g. anthologies published in 2015 are cited 163 times in all SemEval proceedings so far). SA tasks from 2016, SEM from 2014 and IE from 2013 have the highest impact compared within groups (40\%, 28\% and 42\% respectively). One could expect higher numbers of citations for older papers, however, we do not observe this pattern. 

\section{Languages in tasks}
We analysed SemEval it terms of languages used in the tasks (Fig.\ref{fig:languages}). We can distinguish 3 clusters: English-only (except for 3 tasks entirely in Chinese); multi-lingual, which define identical subtasks for several languages; cross-lingual (targeting the use of semantic relation across languages).

In total of 96 tasks, 30 investigated more than one language (multi-lingual and cross-lingual tasks) and 63 tasks were using only English.

The five most popular languages, excluding English are: Spanish (16), French (10), Italian (10), Arabic (8), German (8). Although Chinese is the 1st language in number of speakers, only 4 tasks were organised for Chinese.  

Most of multi-lingual or cross-lingual tasks are related to SA (5 in 2016-2018) or SEM (15 in 2012-2019), and obviously on MT tasks (3 in 2012-2014). There were 3 OT tasks, only one QA task, and no IE tasks. Task 11 in 2017 concerning program synthesis, aiming to translate commands in English into (program) code, attracted only one team.

In 2018 and 2019 the interest of other languages is lower compared to previous years. Languages other than English were proposed only 5 and 3 times, respectively, whereas in 2016 and 2017 we observed the occurrence of respectively 10 and 14 times.

\section{Input and Output Analysis}

In order to better understand the evolution of the semantic complexity of the tasks, we analysed them in terms of the types used to represent input and output data in all subtasks. Based on their descriptions, we devised a list of 25 different abstract types used, then assigning each subtask the most appropriate Input and Output Types.

\subsection{Types and Clusters}
Taking into consideration both their complexity and purpose, we split the type list into 5 clusters: \textit{cluster 1}: document, text, paragraph, sentence, phrase, word, number; \textit{cluster 2}: score, score real value, score whole value, class label, probability distribution; \textit{cluster 3}: entity, attribute, topic, tree, Directed Acyclic Graph (DAG); \textit{cluster 4}: question, answer, query; \textit{cluster 5}: Knowledge Base (KB), program, time interval, timeline, semantic graph, syntactic labeled sentence.

\begin{figure*}
    \centering
    \begin{subfigure}[b]{0.475\textwidth}
        \centering
        \includegraphics[width=0.9\textwidth]{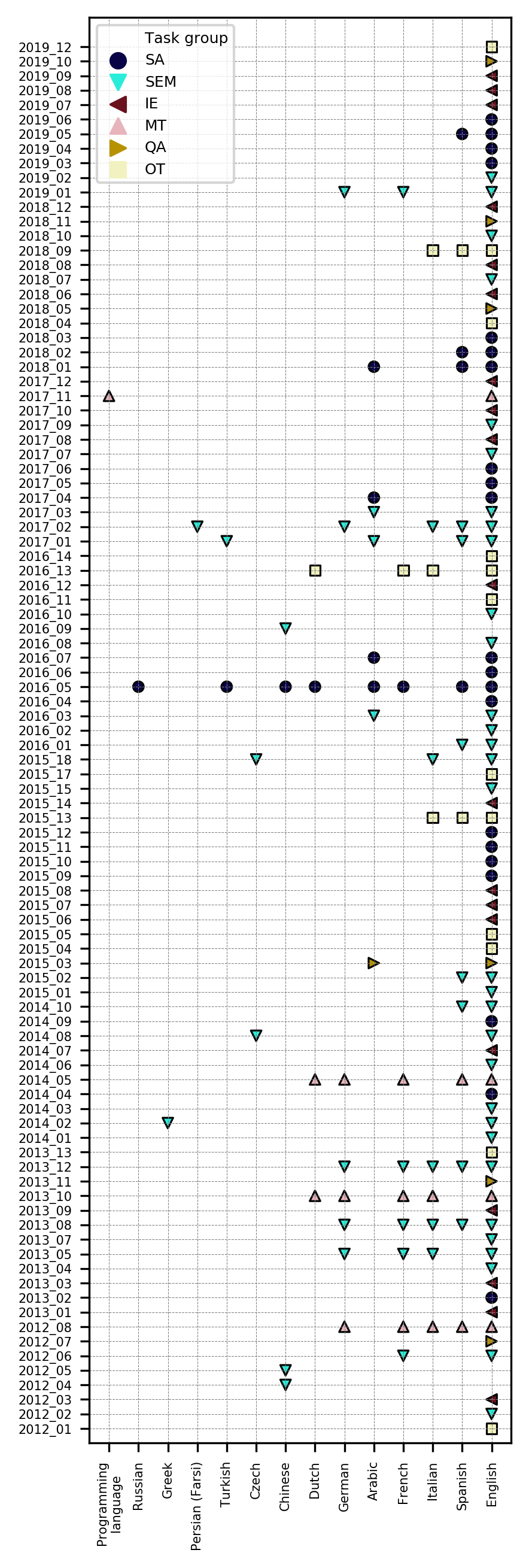}
        \caption[]%
        {{\small Languages used in 96 SemEval tasks from 2012 to 2019}}    
        \label{fig:languages}
    \end{subfigure}
    \hfill
    \begin{subfigure}[b]{0.475\textwidth}  
        \begin{subfigure}[b]{\textwidth}  
        \centering 
        \includegraphics[width=0.9\textwidth]{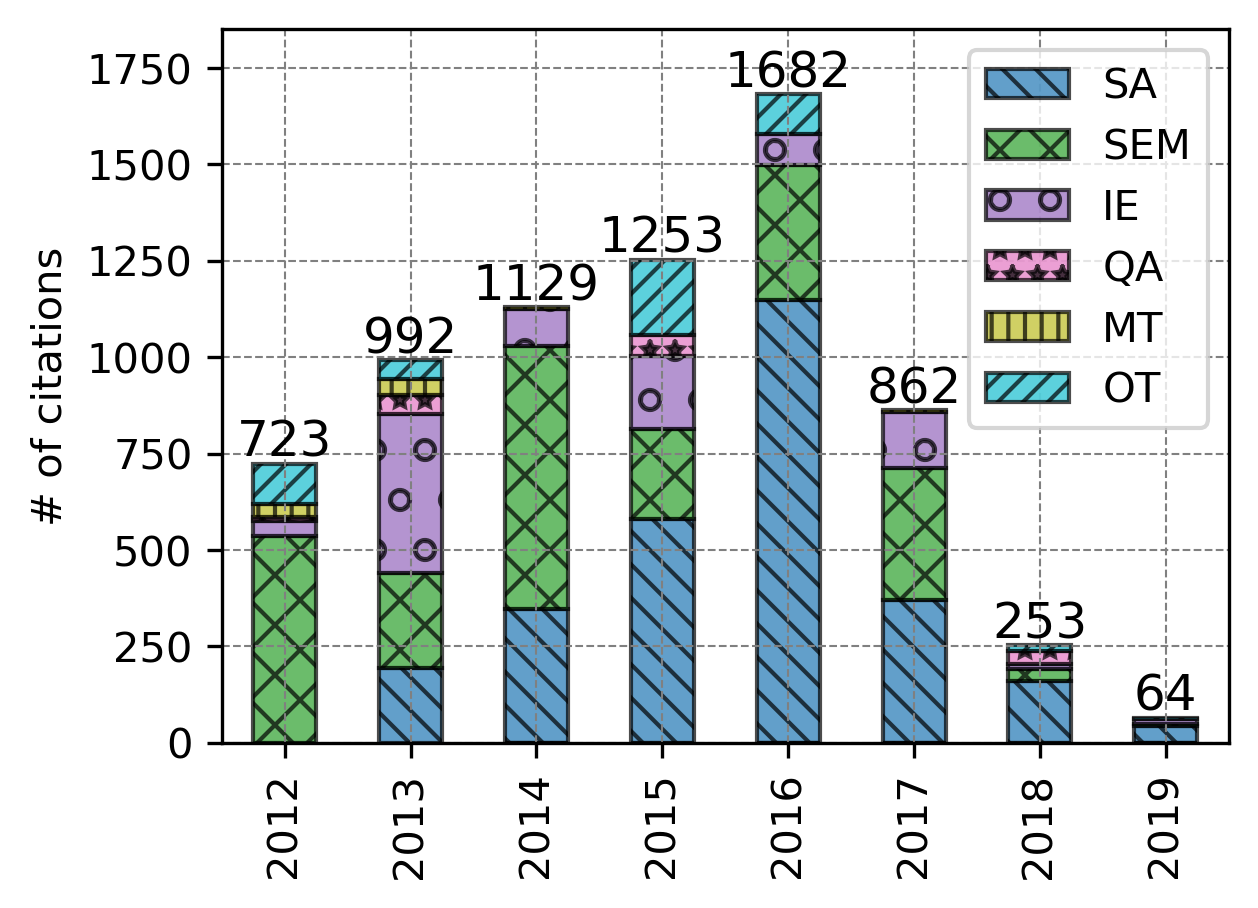}
        \caption[]%
        {{\small Number of task citations published in given year, except for citations in SemEval proceeding}}    
        \label{fig:cit_year}
        \end{subfigure}
        
        \vfill
        \begin{subfigure}[b]{\textwidth}  
        \centering 
        \includegraphics[width=0.9\textwidth]{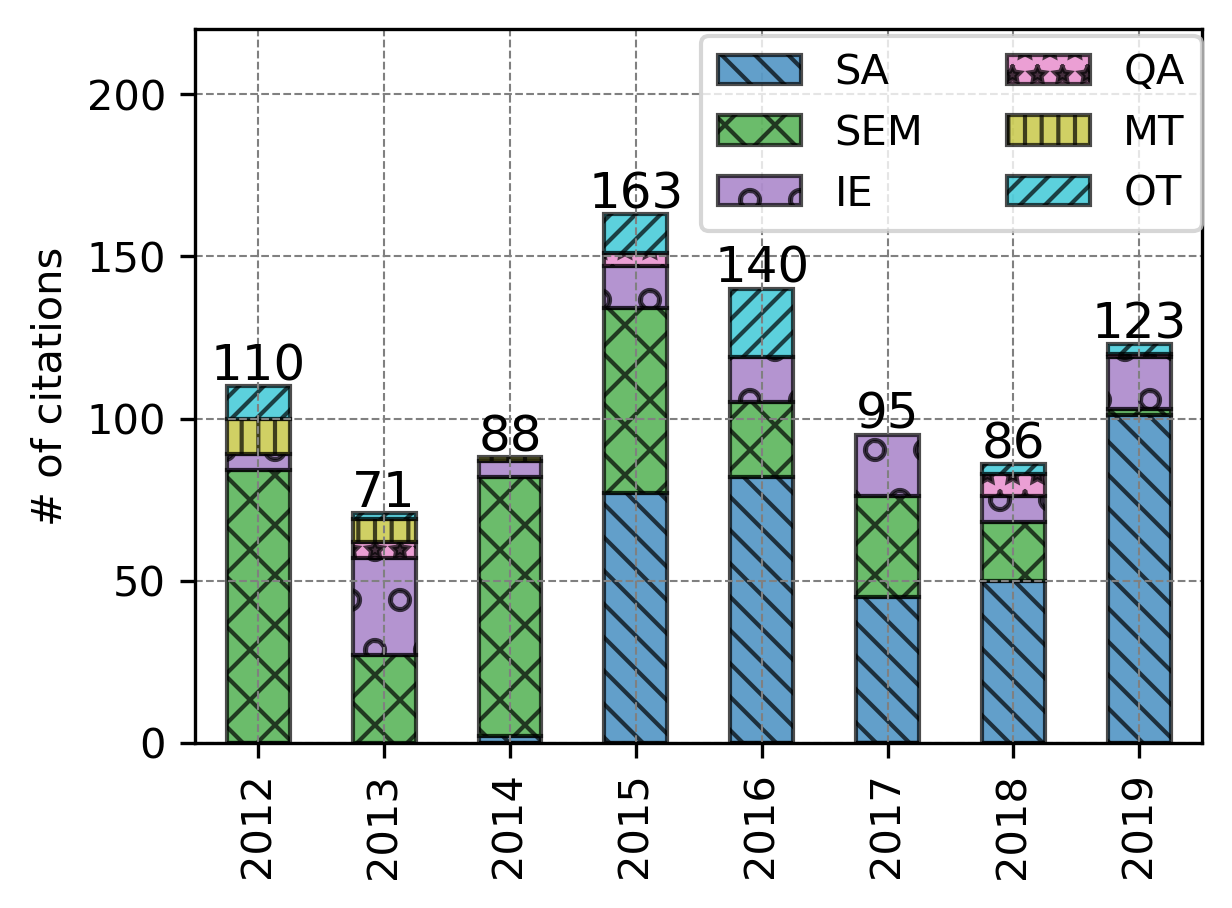}
        \caption[]%
        {{\small Number of task citations from given year in SemEval proceedings}}    
        \label{fig:cit_int}
        \end{subfigure}
        
        \begin{subfigure}[b]{\textwidth}  
        \centering 
        \includegraphics[width=0.9\textwidth]{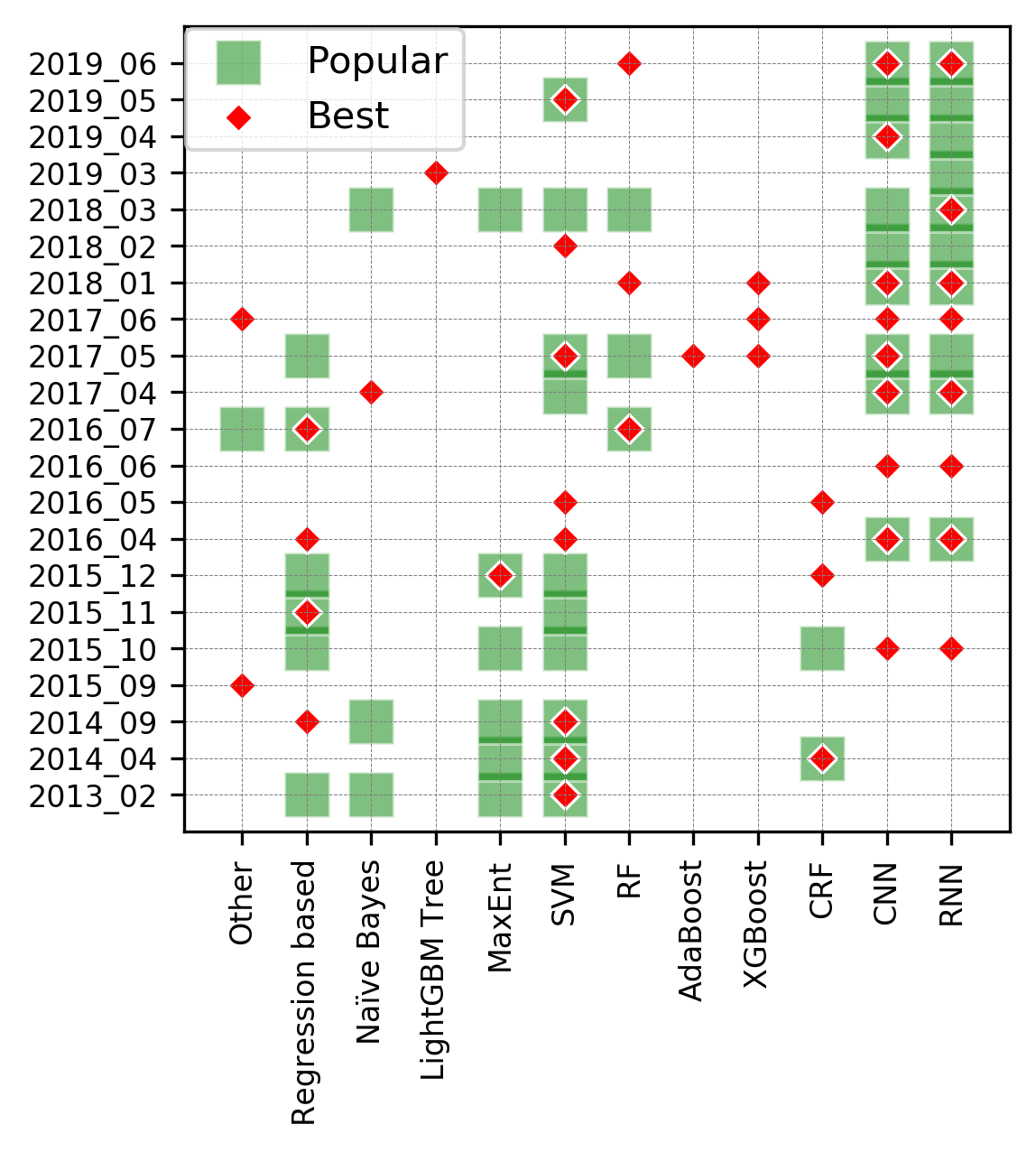}
        \caption[]%
        {{\small Models used in SA tasks from 2012 to 2019 at SemEval}}    
        \label{fig:models}
        \end{subfigure}

    \end{subfigure}
    
    \caption[ AAA ]
    {\small } 
    \label{fig:languages_cit_models}
\end{figure*}

\subsection{Input Types}
As expected, types from \textit{cluster 1} (sequential tokens) make up for 76\% of overall input types used in all tasks (depicted in the Appendix A, Fig.\ref{fig:input}). Most popular input type is paragraph, for which about 60\% of cases represents a tweet. The remaining 24\% is split across \textit{clusters 2, 3, 4 and 5}. A subtle divergence towards the right-hand side can be noticed, starting with 2015, driven mostly by tasks from SA and IE task groups. The most dominant Input Types from each cluster are paragraph, class label, entity, question and KB.

\subsection{Output Types}
\begin{figure*}
    \centering
    \begin{subfigure}[b]{0.475\textwidth}
        \centering
        \includegraphics[height=0.9\textheight]{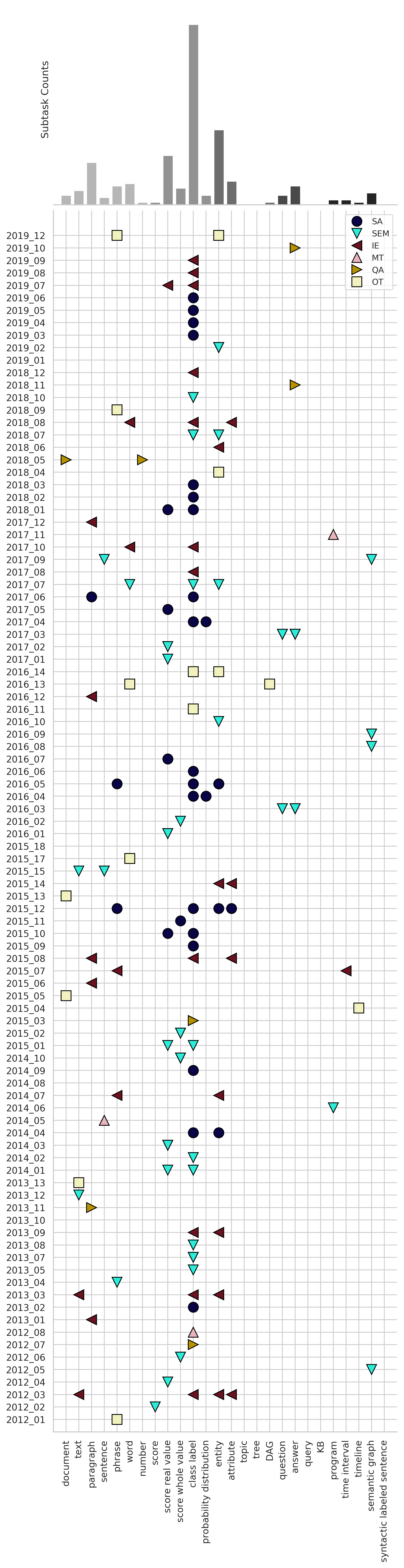}
        \caption[]%
        {{\small Output Types}}    
        \label{fig:output}
    \end{subfigure}
    \hfill
    \begin{subfigure}[b]{0.475\textwidth}  
        \centering 
        \includegraphics[height=0.9\textheight]{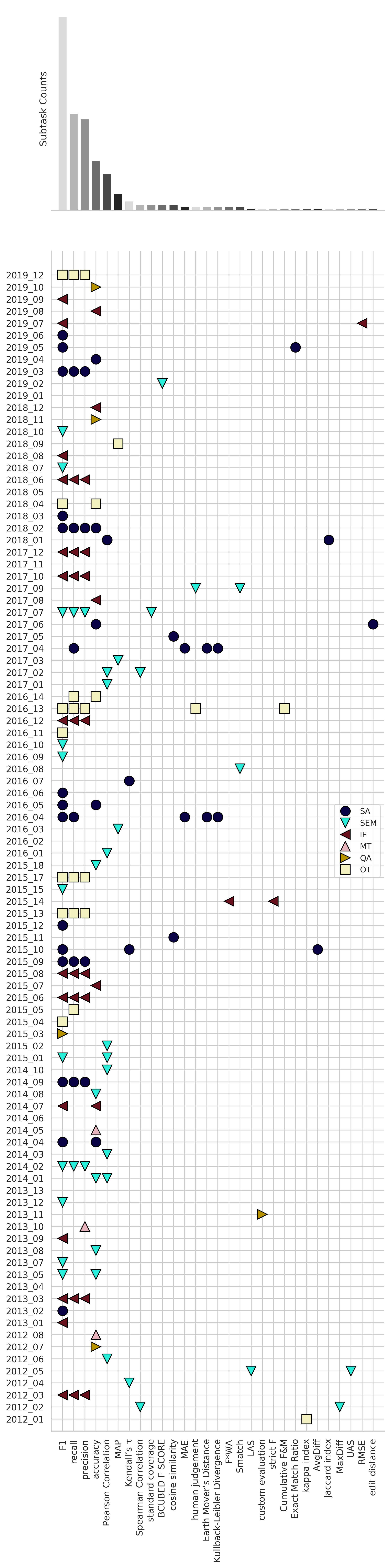}
        \caption[]%
        {{\small Evaluation Metrics}}    
        \label{fig:eval_metrics}
    \end{subfigure}
    
    \caption[ AAA ]
    {\small 96 SemEval tasks from 2012-2019} 
    \label{fig:overall_plots_output_and_eval}
\end{figure*}

As shown in Fig.\ref{fig:output}, data types from \textit{clusters 2 and 3} are the majority in this case, accounting for 68\% of used representations. Class labels are repeatedly employed, especially by SA tasks.
\textit{Cluster 1} types are constantly used across the years, fully dependent on the task types given in a certain year, 78\% of them coming from SEM, IE and OT.
Rarely used are typed from \textit{clusters 4 and 5}, accounting for just 10\% of the total, half of which occur in SEM tasks during 2016 and 2017, complex tasks such as Community Question Answering and Chinese Semantic Dependency Parsing.
We also found a possible relation between output type and popularity. In 2012-2017 tasks where outputs were in \textit{cluster 4 or 5}, attracted 8.3 teams per task on average, while in \textit{clusters 1-3} 13.9 teams/task. However, despite major increase in SemEval popularity, in 2018-2019 the former attracted only 7 teams/task, and the latter 43.5 teams/task.
The group with most type variety is SEM, covering types across all clusters. On the other side of the spectrum, SA has the least variety, despite it being the most popular task group.
The most dominant Output Types from each cluster are paragraph, class label, entity, answer and semantic graph.

\section{Evaluation Metrics}

We counted a total of 29 different evaluation metrics used in SemEval (Fig.\ref{fig:eval_metrics}). 

At a subtask level, the most frequent metric is F1-score, with 105 usages, followed by recall and precision, with 51 and 49 usages respectively, and accuracy, with 26 usages. F1, recall and precision are frequently jointly used, the last two playing the role of supporting break-down metrics for F1 in 95\% of cases. This combination is very popular, especially for IE tasks, almost half of the use coming from this task group.

The top 5 evaluation metrics make up 84\% of the total number of metrics used in all years, last 12 (almost half) being only used once. In 89\% of cases when rare evaluation metrics (from Kendall's T to the right) are used, they occur in SA and SEM tasks e.g. Jaccard index in Affect in Tweets (2018) or Smatch in Meaning Representation Parsing (2016). Furthermore, 67\% of the least used evaluation metrics (only used 3 times or less) appear in 2015-2017, the same period when we could see tasks experimenting the most with input and output types.

\subsection{Evaluation Metrics against Output Types}
F1, recall and precision (depicted in Appendix A, Fig.\ref{fig:heatmap}) are mostly used for output types such as class label, paragraph and entity (each of which is the top output type from their clusters). Meanwhile, for output types represented by score, most used evaluation metrics are Pearson Correlation, Kendall's T, cosine similarity and Spearman Correlation. 
MAP, the 6th most used evaluation metric, is mostly used for ranked questions/answers either in recurring tasks such as Community Question Answering. 
Human judgment was only used twice, in Taxonomy Extraction Evaluation (2016) and Abstract Meaning Representation Parsing and Generation (2017). For further reference, see Appendix A.

\section{Zooming in into Sentiment Analysis}

\subsection{System architectures}
The systematic analysis of the prevalent methods and architectures imposed particular challenges with regard to the data extraction process due to the intrinsic complexity of tasks (many systems include the composition of pre-processing techniques, rules, hand-crafted features and combinations of algorithms). Additionally, for the majority of task description papers, there is no systematic comparison between systems within a task, and consequently within group or years. 

Due to the consistent presence of SA along all years, we present an overview of the evolution of system architectures used in SA from 2013 to 2019 (Fig.\ref{fig:models}). In this analysis we focus on the best performing architectures. More than one best model in a task signifies best models in subtasks or that the final system was an ensemble of several algorithms. \textit{Regression based} model encompasses linear, logistic, or Gaussian regression, and  \textit{Other} includes all rule-based or heavily hand-crafted models.

We observe a drift in popularity of architectures from ML algorithms (2013-2016) to deep learning (DL) models (2017-2019). 

Despite the major adoption of DL models, traditional ML algorithms are consistently in use, both as separate models and as ensembles with DL. This is also true for other types of tasks. In many task description papers from 2018-2019, one can find ML-based systems as top performing participants. SVM-based models are still popular and in some tasks outperforms DL (2018-2, 2019-5).

In the analysis of system architectures one needs to take into account that best system depends not only on the core algorithm but also on the team expertise and supporting feature sets and language resources.


\begin{figure}[b!]
\centering
\includegraphics[width= 0.95\textwidth]{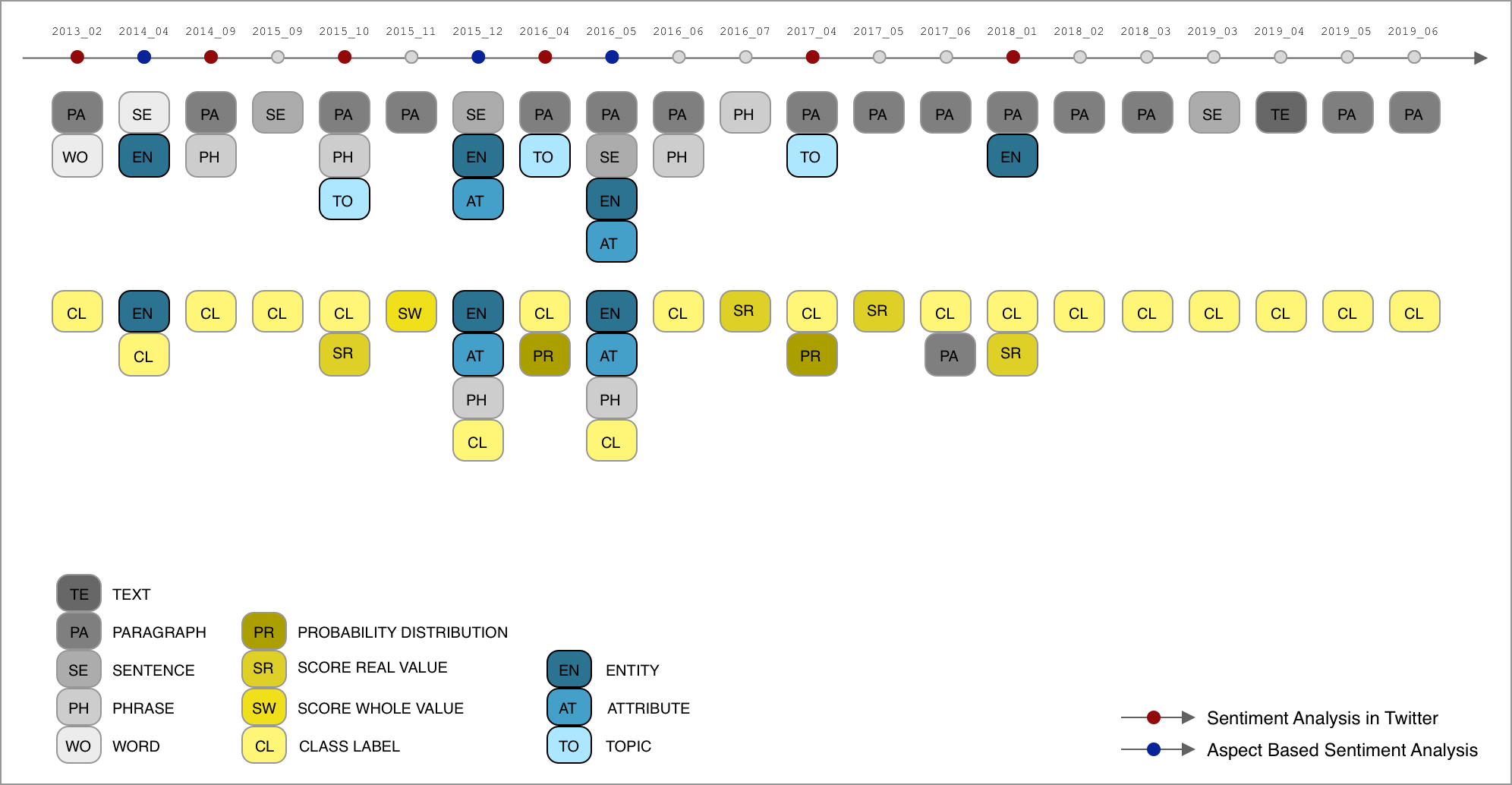}
\caption{Timeline of Input Types (upper row) and Output Types (lower row) in Sentiment Analysis tasks at SemEval 2013-2019.}    
\label{fig:timeline}
\end{figure}

\subsection{Representations}

The output of the SA related tasks provide an account of the evolution of sentiment and emotion representation in this community from 2013 until 2019 (Fig.\ref{fig:timeline}).

At a discrete level, the number of maximum class labels representing sentiment intensity grew from 3 in 2013 to 7 in 2019.
At a continuous score level, real-valued scores associated with sentiment was first used in 2015; in 2016 it switched to sentiment intensity; in 2017 it was being used as a way to determine the intensity of an emotion component out of 11 emotion types (rather than a single one, or the generic emotional intensity of a sentence).
In terms of targeted subject, the tasks grew more granular over time: paragraph/word (2013), aspect terms (2014), sentence topic (2015), person (2016). 
Additionally, discourse evolved from simpler opinionated text in the direction of figurative language, for example: handling irony and metaphor in SA (2015), phrases comparison/ranking in terms of sense of humor (2017), irony detection (2018) and contextual emphasis (2019).

\section{Discussion: What is SemEval evaluating?}

The results of the analysis substantiate the following core claims, which summarises some of the trends identified in this paper: 

\begin{itemize}

\item There is evidence of significant impact of SemEval in the overall NLP community.

\item SemEval contributed to the construction of a large and diverse set of challenges with regard to semantic representation, supporting resources and evaluation methodologies and metrics.

\item SemEval is becoming heavily biased towards solving classification/regression problems. We observe a major interest in tasks where the expected output is a binary or multi-class label or within a continuous real valued score. 

\item Sentiment Analysis tasks accounts for a disproportional attention from the community. 


\item There are two parallel narratives running on SemEval: low entry barrier and state-of-the-art defining. SemEval contains a rich corpus of unaddressed and complex NLP tasks, which are eclipsed by the easier low entry barrier tasks. This points to the double function of SemEval which performs a pedagogical task, serving as an entry point for early career researchers to engage within the NLP community and a state-of-the-art forum for pushing the boundaries of natural language interpretation. With the popularity of NLP applications and Deep Learning, the former function is eclipsing the latter.

\item There is a significant trend to decrease the variety in the output and evaluation metrics in the recent years. While in previous years, tasks focused more on novel and exploratory tasks, recent tasks have explored, probably due to emergence of out-of-the-box DL models, this variety significantly decreased. Consequently, participants focus on easier tasks, which in part dissipates the community potential to address long-term challenges.  

\item Despite the recent interest in neural-based architectures, there is clear evidence of the longevity and lasting impact of older NLP methods.

\end{itemize}

\section{Recommendations}

We believe that this paper can serve as a guideline for the selection and organisation of future SemEval tasks. Based on the analyses performed on this paper, these are the main recommendations:

\begin{itemize}

\item Prioritise tasks which have a clear argument on semantic and methodological challenges and novelty.

\item Differentiate challenges which have a competition/pedagogical purpose from research tasks.

\item Support the systematic capture of task metadata and submission data in a structured manner. This will allow for an efficient comparison between SemEval tasks and deriving insights for future SemEval editions.

\end{itemize}

\section{Conclusions}

This paper reported a systematic quantitative analysis of SemEval, one of the primary venues for the empirical evaluation of NLP systems. The analysis, which provides a detailed breakdown of 96 tasks in the period between 2012-2019, provided quantitative evidence that: (i) SemEval has a significant impact in the overall NLP community, (ii) there is a recent drift towards a bias in the direction of Deep Learning classification methods which is eclipsing the research function of SemEval and (iii) that there is longevity and impact of older NLP methods in comparison to Deep Learning methods.

\bibliographystyle{coling}
\bibliography{refs}

\begin{thebibliography}{}

\bibitem[\protect\citename{Agirre \bgroup et al.\egroup
  }2012]{semeval-2012-sem}
Eneko Agirre, Johan Bos, Mona Diab, Suresh Manandhar, Yuval Marton, and Deniz
  Yuret, editors.
\newblock 2012.
\newblock {\em *{SEM} 2012: The First Joint Conference on Lexical and
  Computational Semantics {--} Volume 1: Proceedings of the main conference and
  the shared task, and Volume 2: Proceedings of the Sixth International
  Workshop on Semantic Evaluation ({S}em{E}val 2012)}, Montr{\'e}al, Canada,
  7-8 June. Association for Computational Linguistics.

\bibitem[\protect\citename{Apidianaki \bgroup et al.\egroup
  }2018]{semeval-2018-international}
Marianna Apidianaki, Saif~M. Mohammad, Jonathan May, Ekaterina Shutova, Steven
  Bethard, and Marine Carpuat, editors.
\newblock 2018.
\newblock {\em Proceedings of The 12th International Workshop on Semantic
  Evaluation}, New Orleans, Louisiana, June. Association for Computational
  Linguistics.

\bibitem[\protect\citename{Barrault \bgroup et al.\egroup
  }2019]{barrault-EtAl:2019:WMT}
Loïc Barrault, Ondřej Bojar, Marta~R. Costa-jussà, Christian Federmann, Mark
  Fishel, Yvette Graham, Barry Haddow, Matthias Huck, Philipp Koehn, Shervin
  Malmasi, Christof Monz, Mathias Müller, Santanu Pal, Matt Post, and Marcos
  Zampieri.
\newblock 2019.
\newblock Findings of the 2019 conference on machine translation (wmt19).
\newblock In {\em Proceedings of the Fourth Conference on Machine Translation
  (Volume 2: Shared Task Papers, Day 1)}, pages 1--61, Florence, Italy, August.
  Association for Computational Linguistics.

\bibitem[\protect\citename{Bethard \bgroup et al.\egroup
  }2016]{semeval-2016-international}
Steven Bethard, Marine Carpuat, Daniel Cer, David Jurgens, Preslav Nakov, and
  Torsten Zesch, editors.
\newblock 2016.
\newblock {\em Proceedings of the 10th International Workshop on Semantic
  Evaluation ({S}em{E}val-2016)}, San Diego, California, June. Association for
  Computational Linguistics.

\bibitem[\protect\citename{Bethard \bgroup et al.\egroup
  }2017]{semeval-2017-international}
Steven Bethard, Marine Carpuat, Marianna Apidianaki, Saif~M. Mohammad, Daniel
  Cer, and David Jurgens, editors.
\newblock 2017.
\newblock {\em Proceedings of the 11th International Workshop on Semantic
  Evaluation ({S}em{E}val-2017)}, Vancouver, Canada, August. Association for
  Computational Linguistics.

\bibitem[\protect\citename{Chapman \bgroup et al.\egroup
  }2011]{10.1136/amiajnl-2011-000465}
Wendy~W Chapman, Prakash~M Nadkarni, Lynette Hirschman, Leonard~W D'Avolio,
  Guergana~K Savova, and Ozlem Uzuner.
\newblock 2011.
\newblock {Overcoming barriers to NLP for clinical text: the role of shared
  tasks and the need for additional creative solutions}.
\newblock {\em Journal of the American Medical Informatics Association},
  18(5):540--543, 09.

\bibitem[\protect\citename{Filannino and Uzuner}2018]{PMID:30157522}
Michele Filannino and Özlem Uzuner.
\newblock 2018.
\newblock Advancing the state of the art in clinical natural language
  processing through shared tasks.
\newblock {\em Yearbook of medical informatics}, 27(1):184—192, August.

\bibitem[\protect\citename{Harzing}2007]{publish-and-perish}
A.W. Harzing.
\newblock 2007.
\newblock Publish or perish.
\newblock available from https://harzing.com/resources/publish-or-perish.

\bibitem[\protect\citename{Jimenez \bgroup et al.\egroup }2015]{JIMENEZ2015}
Sergio Jimenez, Fabio~A. Gonzalez, and Alexander Gelbukh.
\newblock 2015.
\newblock {Soft Cardinality in Semantic Text Processing: Experience of the
  SemEval International Competitions}.
\newblock {\em {Polibits}}, pages 63 -- 72, 06.

\bibitem[\protect\citename{Manandhar and
  Yuret}2013]{semeval-2013-joint-lexical}
Suresh Manandhar and Deniz Yuret, editors.
\newblock 2013.
\newblock {\em Second Joint Conference on Lexical and Computational Semantics
  (*{SEM}), Volume 2: Proceedings of the Seventh International Workshop on
  Semantic Evaluation ({S}em{E}val 2013)}, Atlanta, Georgia, USA, June.
  Association for Computational Linguistics.

\bibitem[\protect\citename{May \bgroup et al.\egroup
  }2019]{semeval-2019-international}
Jonathan May, Ekaterina Shutova, Aurelie Herbelot, Xiaodan Zhu, Marianna
  Apidianaki, and Saif~M. Mohammad, editors.
\newblock 2019.
\newblock {\em Proceedings of the 13th International Workshop on Semantic
  Evaluation}, Minneapolis, Minnesota, USA, June. Association for Computational
  Linguistics.

\bibitem[\protect\citename{Nakov and Zesch}2014]{semeval-2014-international}
Preslav Nakov and Torsten Zesch, editors.
\newblock 2014.
\newblock {\em Proceedings of the 8th International Workshop on Semantic
  Evaluation ({S}em{E}val 2014)}, Dublin, Ireland, August. Association for
  Computational Linguistics.

\bibitem[\protect\citename{Nakov \bgroup et al.\egroup
  }2015]{semeval-2015-international}
Preslav Nakov, Torsten Zesch, Daniel Cer, and David Jurgens, editors.
\newblock 2015.
\newblock {\em Proceedings of the 9th International Workshop on Semantic
  Evaluation ({S}em{E}val 2015)}, Denver, Colorado, June. Association for
  Computational Linguistics.

\bibitem[\protect\citename{Nakov \bgroup et al.\egroup }2016]{Nakov2016}
Preslav Nakov, Sara Rosenthal, Svetlana Kiritchenko, Saif~M. Mohammad, Zornitsa
  Kozareva, Alan Ritter, Veselin Stoyanov, and Xiaodan Zhu.
\newblock 2016.
\newblock Developing a successful semeval task in sentiment analysis of twitter
  and other social media texts.
\newblock {\em Language Resources and Evaluation}, 50(1):35--65, Mar.

\bibitem[\protect\citename{Nissim \bgroup et al.\egroup
  }2017]{nissim-etal-2017-last}
Malvina Nissim, Lasha Abzianidze, Kilian Evang, Rob van~der Goot, Hessel
  Haagsma, Barbara Plank, and Martijn Wieling.
\newblock 2017.
\newblock Last words: Sharing is caring: The future of shared tasks.
\newblock {\em Computational Linguistics}, 43(4):897--904, December.

\bibitem[\protect\citename{Parra~Escart{\'\i}n \bgroup et al.\egroup
  }2017]{parra-escartin-etal-2017-ethical}
Carla Parra~Escart{\'\i}n, Wessel Reijers, Teresa Lynn, Joss Moorkens, Andy
  Way, and Chao-Hong Liu.
\newblock 2017.
\newblock Ethical considerations in {NLP} shared tasks.
\newblock In {\em Proceedings of the First {ACL} Workshop on Ethics in Natural
  Language Processing}, pages 66--73, Valencia, Spain, April. Association for
  Computational Linguistics.

\bibitem[\protect\citename{Sygkounas \bgroup et al.\egroup
  }2016]{Sygkounas2016}
Efstratios Sygkounas, Giuseppe Rizzo, and Rapha{\"e}l Troncy.
\newblock 2016.
\newblock A replication study of the top performing systems in semeval twitter
  sentiment analysis.
\newblock In Paul Groth, Elena Simperl, Alasdair Gray, Marta Sabou, Markus
  Kr{\"o}tzsch, Freddy Lecue, Fabian Fl{\"o}ck, and Yolanda Gil, editors, {\em
  The Semantic Web -- ISWC 2016}, pages 204--219, Cham. Springer International
  Publishing.

\end{thebibliography}

  

\appendix
\counterwithin{figure}{section}
\section{Appendix}

\begin{figure}[ht]
\includegraphics[width = \linewidth]{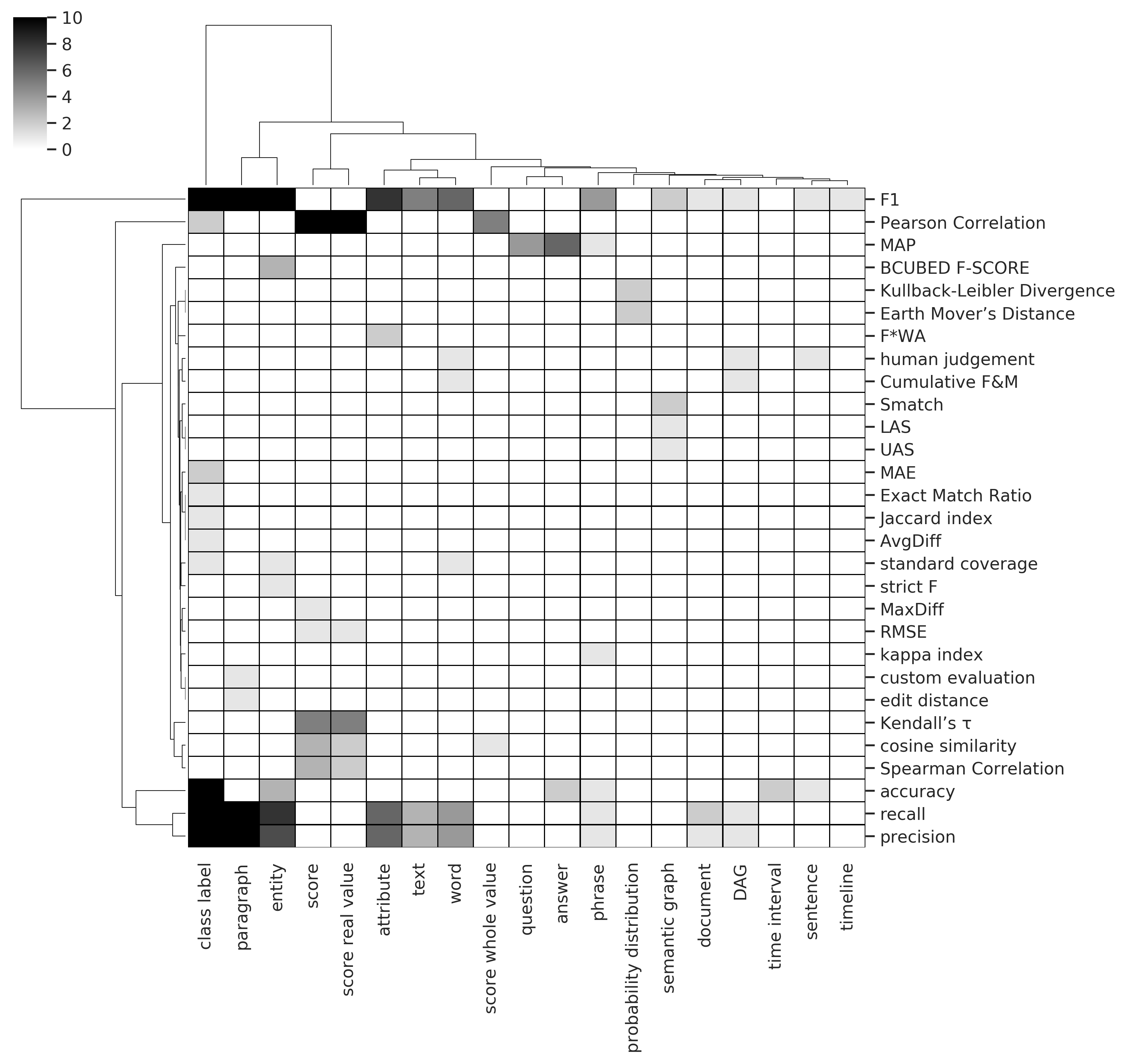}
\centering
\caption{Heatmap for the Evaluation Metrics and Output Types}
\label{fig:heatmap}
\end{figure}

\begin{figure}[b!]
\includegraphics[height= 0.95\textheight]{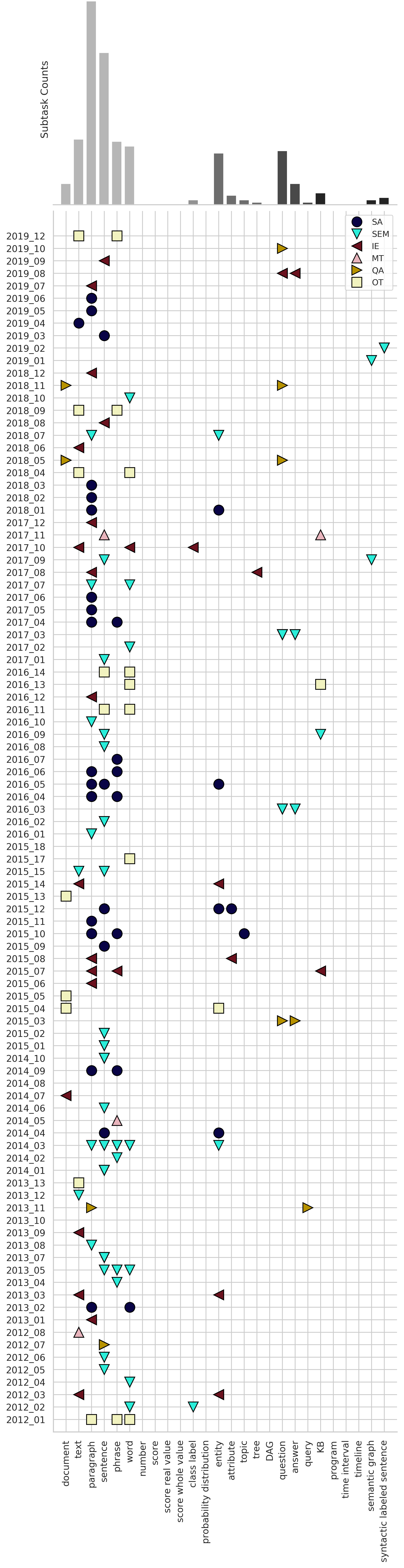}
\centering
\caption{Input Types used in SemEval tasks from 2012 to 2019}
\label{fig:input}
\end{figure}

\end{document}